\documentclass[]{onethree}

\usepackage[toc,page,header]{appendix}
\usepackage{minitoc}
\usepackage{amsmath,amssymb}
\usepackage{bm}
\usepackage{wrapfig}
\usepackage{enumitem}
\usepackage[utf8]{inputenc} 
\usepackage[T1]{fontenc}    
\usepackage{hyperref}       
\usepackage{url}            
\usepackage{booktabs}       
\usepackage{amsfonts}       
\usepackage{nicefrac}       
\usepackage{microtype}      
\usepackage{amsmath}         
\usepackage{graphicx}
\usepackage{multirow} 
\PassOptionsToPackage{table}{xcolor}
\usepackage[table]{xcolor}
\usepackage{booktabs}
\usepackage{adjustbox}
\usepackage{array}
\usepackage{pifont}
\usepackage{makecell}
\usepackage{caption}
\newcommand{\gain}[1]{#1\%\textcolor{green}{\(\uparrow\)}}
\newcommand{\drop}[1]{#1\%\textcolor{red}{\(\downarrow\)}}

\title{CitySTAR: Structured and Topology-Aware Reasoning for Open-Vocabulary Urban 3D Grounding}

\renewcommand{\authorlist}{%
    \authorformat[1,*]{Shuai~Zhang}\quad
    \authorformat[1,*]{Hongye~Hou}\quad
    \authorformat[1]{Qinghe~Liu}\quad
    \authorformat[1]{Zhuoxiao~Li}\\[2pt]
    \authorformat[1]{Dongli~Wu}\quad
    \authorformat[1]{Jing~Ou}\quad
    \authorformat[2]{Yuan~Liu}\quad
    \authorformat[1,\dagger]{Wufan~Zhao}%
}

\makeatletter
\patchcmd{\mymaketitle}
  {\affiliationlist\par \vskip 3mm}
  {\affiliationlist\par \vskip 0mm}
  {}
  {\PackageWarning{citystar-layout}{Affiliation spacing patch failed}}
\makeatother
\affiliation[1]{HKUST(GZ)}
\affiliation[2]{HKUST}

\abstract{
3D grounding aims to localize target entities in complex scenes from natural language and plays a fundamental role in embodied perception and spatial reasoning. However, existing approaches mostly rely on feature similarity or direct matching, making it difficult to connect natural-language intent with the implicit semantic and geometric structures hidden in billion-scale urban point clouds. We reformulate city-scale 3D grounding as structured constraint reasoning, where description semantics are organized into computable cross-modal constraints over open-vocabulary 3D entities, attributes, and spatial relations. We present \textbf{CitySTAR}, a training-free framework for reasoning-driven urban 3D grounding. CitySTAR lifts raw billion-scale urban point clouds into a query-ready scene graph of open-vocabulary 3D instances, with CodeLLM-driven tools supplying multimodal evidence for node attributes and 3D spatial relations. It then models target-context topology with paired hypergraphs and performs bidirectional topology verification for structural disambiguation. Finally, a Reflective Cross-modal Grounding module integrates topology consistency and candidate-centered 2D visual evidence to make decisions over a metric-aware 3D context graph. To further support this setting, we introduce \textbf{CitySTAR-3D}, an enhanced benchmark that improves semantic coverage, instance completeness, bounding-box fidelity, and spatial-relation complexity in city-scale 3D grounding. Extensive experiments show that CitySTAR consistently improves open-world urban 3D grounding while maintaining strong interpretability and generalization.
}

\usepackage{etoolbox}

\makeatletter

\patchcmd{\mymaketitle}
  {\vspace*{0.65cm}}
  {}
  {}
  {\PackageWarning{citystar-layout}{Empty resource space patch failed}}

\patchcmd{\mymaketitle}
  {\begin{tcolorbox}}
    {\begin{tcolorbox}[
        left=5mm,
        right=5mm,
        top=4mm,
        bottom=4mm,
        boxsep=0pt,
        after skip=0pt
    ]}
  {}
  {\PackageWarning{citystar-layout}{Abstract padding patch failed}}

\patchcmd{\maketitle}
  {\vskip 8mm}
  {\vskip 2mm}
  {}
  {\PackageWarning{citystar-layout}{Title spacing patch failed}}

\makeatother

\renewcommand{\abstractinfont}{\fontsize{10.5}{13}\selectfont}

\newsavebox{\citytitlebox}
\newsavebox{\citycaptionbox}
\newlength{\cityimageheight}
\makeatletter
\patchcmd{\mymaketitle}
  {\tcbset{enhanced,frame hidden}}
  {\vspace{-5mm}\tcbset{enhanced,frame hidden}}
  {}
  {\PackageWarning{citystar-layout}{Abstract outer spacing patch failed}}
\makeatother
\begin{document}

\begin{lrbox}{\citytitlebox}
    \begin{minipage}{\textwidth}
        \maketitle
    \end{minipage}
\end{lrbox}

\begin{lrbox}{\citycaptionbox}
    \begin{minipage}{\textwidth}
        \captionsetup{
            font=small,
            skip=0pt,
            belowskip=0pt
        }
        
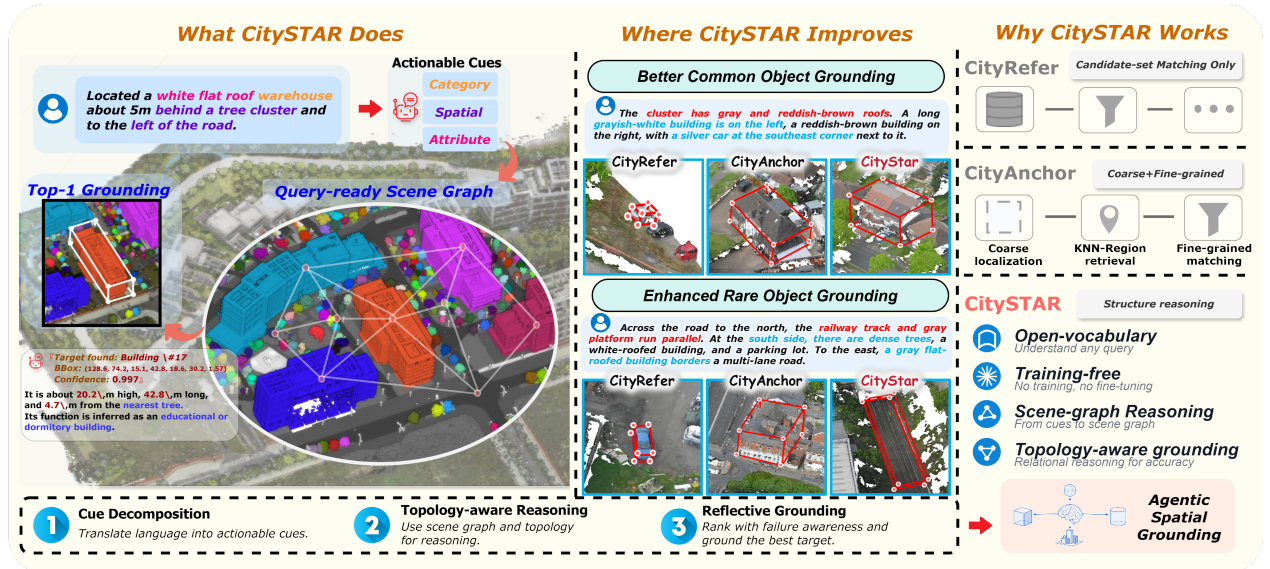
\captionof{figure}{
        CitySTAR upgrades city-scale 3D grounding from candidate matching
        to structured reasoning.
        It combines actionable cue organization, query-ready scene graphs,
        and metric-aware topology-aware grounding to deliver better target
        disambiguation and more accurate localization.
        }
        \label{fig:teaser}
    \end{minipage}
\end{lrbox}

\setlength{\cityimageheight}{%
    \dimexpr
        \textheight
        -\ht\citytitlebox
        -\dp\citytitlebox
        -\ht\citycaptionbox
        -\dp\citycaptionbox
        -12pt
    \relax
}

\thispagestyle{firststyle}

\par
\nointerlineskip
\vbox to \textheight{%
    \offinterlineskip

    \hbox{\usebox{\citytitlebox}}%

    \vskip 6pt
    \vfil

    \hbox to \textwidth{%
        \hfil
        \includegraphics[
            width=\textwidth,
        ]{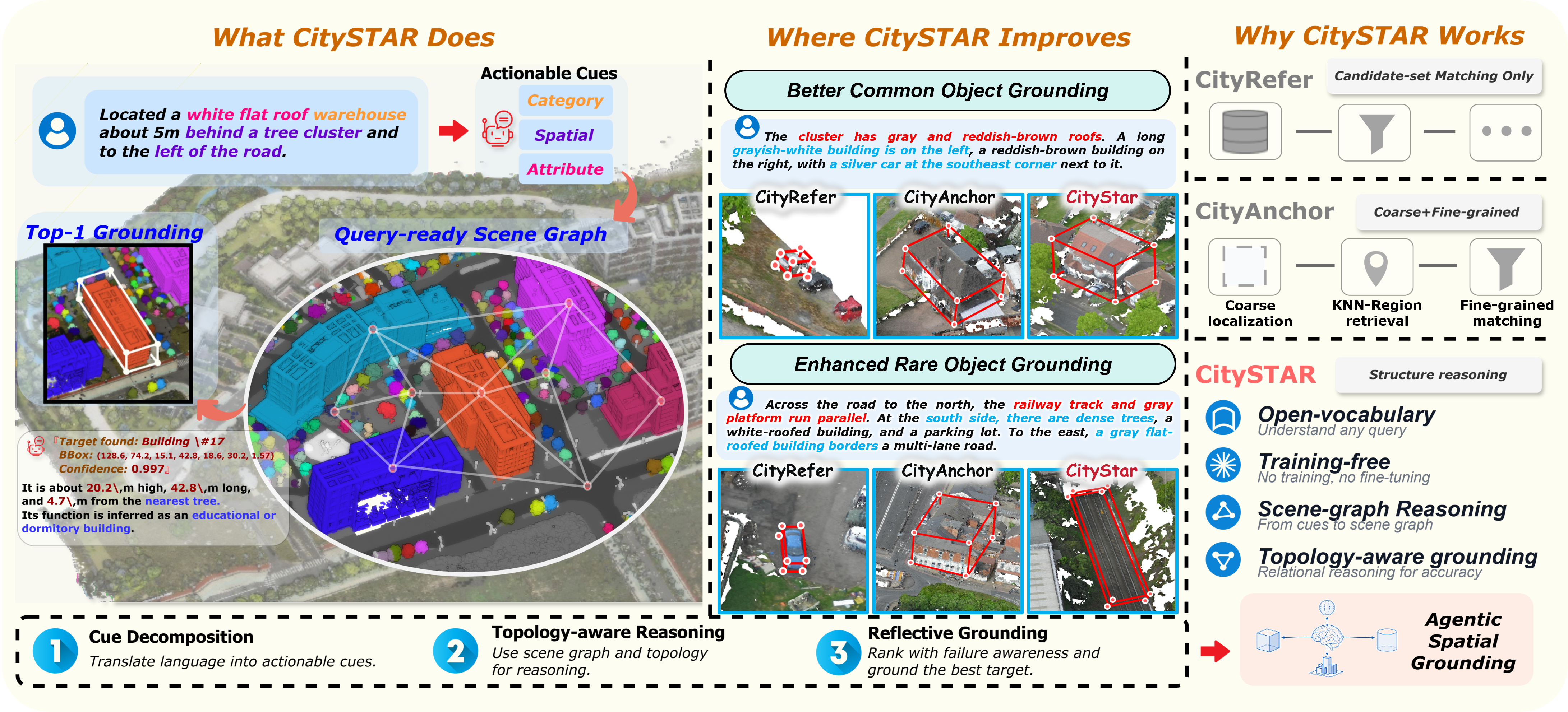}%
        \hfil
    }%

    \vfil
    \vskip 6pt

    \hbox{\usebox{\citycaptionbox}}%
    \kern 0pt
}

\clearpage






\section{Introduction}
\label{intro}

3D visual grounding aims to localize target entities in a 3D scene according to natural-language descriptions, serving as a key interface between language understanding and spatial perception~\citep{chen2020scanrefer}. While this task has been widely studied in indoor environments, extending it to city-scale scenes introduces a substantially different problem. Urban point clouds often contain billion-scale points and repeated object patterns. Incomplete observations and dense similar-looking entities further make city-scale grounding a relational reasoning problem rather than simple object retrieval. A query such as \textit{“the white flat-roof warehouse behind the tree cluster and to the left of the road”} requires the model to identify the target category, recognize visual attributes, locate contextual objects, and verify spatial topology in a large urban context. This makes city-scale 3D grounding important for urban digital twins, embodied navigation, and large-scale spatial reasoning~\citep{liu2025survey,xie2025generative,zhu2025move,chen2026spatialllm}.

Existing grounding methods have made steady progress in 2D vision~\citep{deng2021transvg,zhan2023rsvg,cao2024emergent}, indoor 3D scenes~\citep{liu2026view,li2025seeground,huang2025viewsrd,liu2025reasongrounder}, and recent city-scale settings~\citep{miyanishi2023cityrefer,li2025cityanchor}. However, most methods still follow a matching-centered paradigm, where language is aligned with object features, candidate regions, or learned embeddings. CityRefer~\citep{miyanishi2023cityrefer} selects targets from a predefined candidate set through joint embedding and similarity matching. CityAnchor~\citep{li2025cityanchor} improves the search process with a coarse-to-fine localization strategy. Their semantic spaces remain constrained by predefined category sets, where CityRefer mainly covers four target categories and CityAnchor extends the scope to nine categories. This closed-set design restricts grounding to a limited portion of urban entities and makes it difficult to handle fine-grained or functional objects including warehouses, civic buildings, parking facilities, and tree clusters. These methods are effective when the target can be identified by a known category or simple appearance cue. However, they become fragile when the query involves open-vocabulary entities and relations among multiple objects. In large urban scenes, many candidates may appear visually plausible while still violating the intended target-context topology.

This limitation reflects a deeper issue in current formulations. Spatial relations such as \textit{behind}, \textit{left of}, \textit{adjacent to}, and \textit{between} are often hidden in global embeddings or coarse spatial priors. Existing KNN aggregation struggles to support topological relationships in descriptions. This implicit modeling becomes difficult to scale in billion-point urban scenes with repeated objects and ambiguous local layouts. Attribute-level evidence can narrow the candidate space, but it cannot reliably decide which candidate satisfies the full relational description. The decisive cue often lies not in what the target looks like, but in how it is connected to surrounding entities. Therefore, city-scale 3D grounding requires a mechanism that can organize language, ground intermediate evidence, verify relations, and revise uncertain decisions in a structured manner.

This motivates our view that city-scale 3D grounding should be formulated as \textbf{graph-based chain-of-thought (Cot) reasoning}. A complex grounding query is rarely solved by one-step feature matching, since it usually involves a sequence of intermediate reasoning states. The model extracts actionable cues from language, grounds them onto open-vocabulary urban instances, verifies whether the candidate and contextual entities satisfy the described topology, and makes the final decision by combining multimodal evidence with spatial consistency. Unlike free-form textual CoT, the reasoning chain in 3D grounding should be grounded in explicit scene structures and metric perception directly quantified in 3D space. Each intermediate step should correspond to a computable operation over instance nodes, semantic attributes, metric relations, and local topology. This makes graph-based CoT a suitable form of reasoning for city-scale grounding, because the reasoning process can be represented and verified through scene graphs, paired hypergraphs, and metric-aware 3D context graphs.

Based on this insight, we propose \textbf{CitySTAR}, a training-free framework for city-scale 3D grounding. To the best of our knowledge, CitySTAR is the first attempt to instantiate graph-based CoT-style reasoning for this task. The framework is designed to address three key difficulties in large urban scenes. First, CitySTAR constructs a query-ready scene graph from open-vocabulary 3D instances and uses CodeLLM-driven tool orchestration to ground category, attribute, and spatial cues into a high-recall candidate pool. This avoids exhaustive matching over raw point clouds and supports open-world candidate discovery. Second, CitySTAR represents both language descriptions and candidate neighborhoods as paired hypergraphs, enabling bidirectional topology verification between the target and contextual entities. This turns spatial language into explicit graph constraints and filters candidates that look plausible but violate the described relation. Third, CitySTAR performs candidate-centered 2D visual verification with a visual-text reranker. The visual evidence is projected back onto the graph to support entity nodes and relation edges, allowing the final decision to combine topology consistency with cross-modal appearance evidence. In this way, grounding is transformed from a direct similarity search into a progressive process of cue grounding, topology verification, and graph-supported decision making.

We further introduce \textbf{CitySTAR-3D}, an enhanced benchmark for reasoning-oriented city-scale 3D grounding. Compared with existing datasets, CitySTAR-3D provides broader semantic coverage, more complete instances, clearer bounding boxes, and more complex descriptions involving spatial relations. These properties make it more suitable for evaluating whether a grounding method truly understands target-context topology rather than only matching object appearance. Experiments on CityRefer, CityAnchor, and CitySTAR-3D show that CitySTAR achieves more reliable open-world urban grounding, especially for queries involving complex attributes and spatial relations.

Our contributions are summarized as follows:
\begin{itemize}
    \item We make the first attempt to introduce graph-based CoT-style reasoning into city-scale 3D grounding, reformulating the task as explicit constraint reasoning over open-vocabulary urban instances.
    
    \item We propose \textbf{CitySTAR}, a training-free framework that addresses large-scale search, relation ambiguity, and final decision reliability through CodeLLM-driven cue grounding, paired hypergraph topology verification, and candidate-centered 2D visual verification.
    
    \item We build \textbf{CitySTAR-3D}, an enhanced benchmark that expands semantic categories, improves instance completeness and bounding-box fidelity, and provides more challenging spatial-relation descriptions.
\end{itemize}


\section{Related works}\label{RW}

\textbf{Vision-language models for 3D scene understanding}. Vision-Language Models (VLMs) have advanced 3D scene understanding by improving language-scene alignment and spatial reasoning. In indoor scenes, prior studies~\citep{hong20233d, chen2024ll3da, fu2024scene, yin2023lamm, yang2024llm, tang2024minigpt, li20243dmit, zhang2024agent3d, gu2024conceptgraphs, guo2023point} have shown promising results in fine-grained localization, multimodal interaction, and language-guided scene interpretation. Recent work has extended these capabilities to outdoor and city-scale settings~\citep{chen2024chat3d, sun2025city, chen2026spatialllm, chen20263dcity}. However, existing VLMs are not inherently designed to search over massive 3D spaces or to understand metric 3D geometry and spatial structure directly.

\textbf{3D visual grounding in indoor scenes.} Early 3D visual grounding research focused on indoor target localization. ScanRefer~\citep{chen2020scanrefer} established the task on large-scale 3D scenes based on ScanNet~\citep{dai2017scannet}, while ReferIt3D~\citep{achlioptas2020referit3d} further emphasized instance-level discrimination under semantic ambiguity. Later studies improved grounding through spatial relation modeling and contextual reasoning~\citep{roh2022languagerefer, yang2023exploiting, gu2024conceptgraphs, liu2026view, liu2025reasongrounder}, multi-view observations for cross-view association~\citep{gu2024conceptgraphs, huang2022multi, yang2021sat, zhang2024agent3d, huang2025viewsrd}, and the integration of large language models~\citep{hong20233d, chen2024ll3da, fu2024scene, li2025seeground}. 
When transferred to outdoor urban scenes, these methods face a sharply enlarged candidate space and often lack the end-to-end encoding capacity needed for raw, large-scale 3D environments.

\textbf{3D visual grounding in city-scale environments.} 3D visual grounding has recently moved from indoor scenes to large-scale urban environments. TouchDown~\citep{chen2019touchdown} explored language-guided spatial reasoning and localization in real-world outdoor settings. Text2Pos~\citep{kolmet2022text2pos} further introduced language-driven 3D point cloud localization with the KITTI360Pose benchmark. CityRefer~\citep{miyanishi2023cityrefer} established the first city-scale 3D visual grounding benchmark and highlighted the ambiguity of dense urban landmarks. More recently, CityAnchor~\citep{li2025cityanchor} improved grounding through a self-annotated synthetic dataset and a coarse-to-fine multimodal reasoning strategy. 3EED~\citep{li20253eed} further extends outdoor 3D grounding to multi-platform RGB-LiDAR settings across vehicles, drones, and quadrupeds. 
Nevertheless, city-scale grounding remains challenging because it requires precise open-vocabulary object localization, metric-aware 3D perception, and topology reasoning over target-context relations among urban entities.
\begin{figure}[h]
    \centering
    \includegraphics[width=0.9\linewidth]{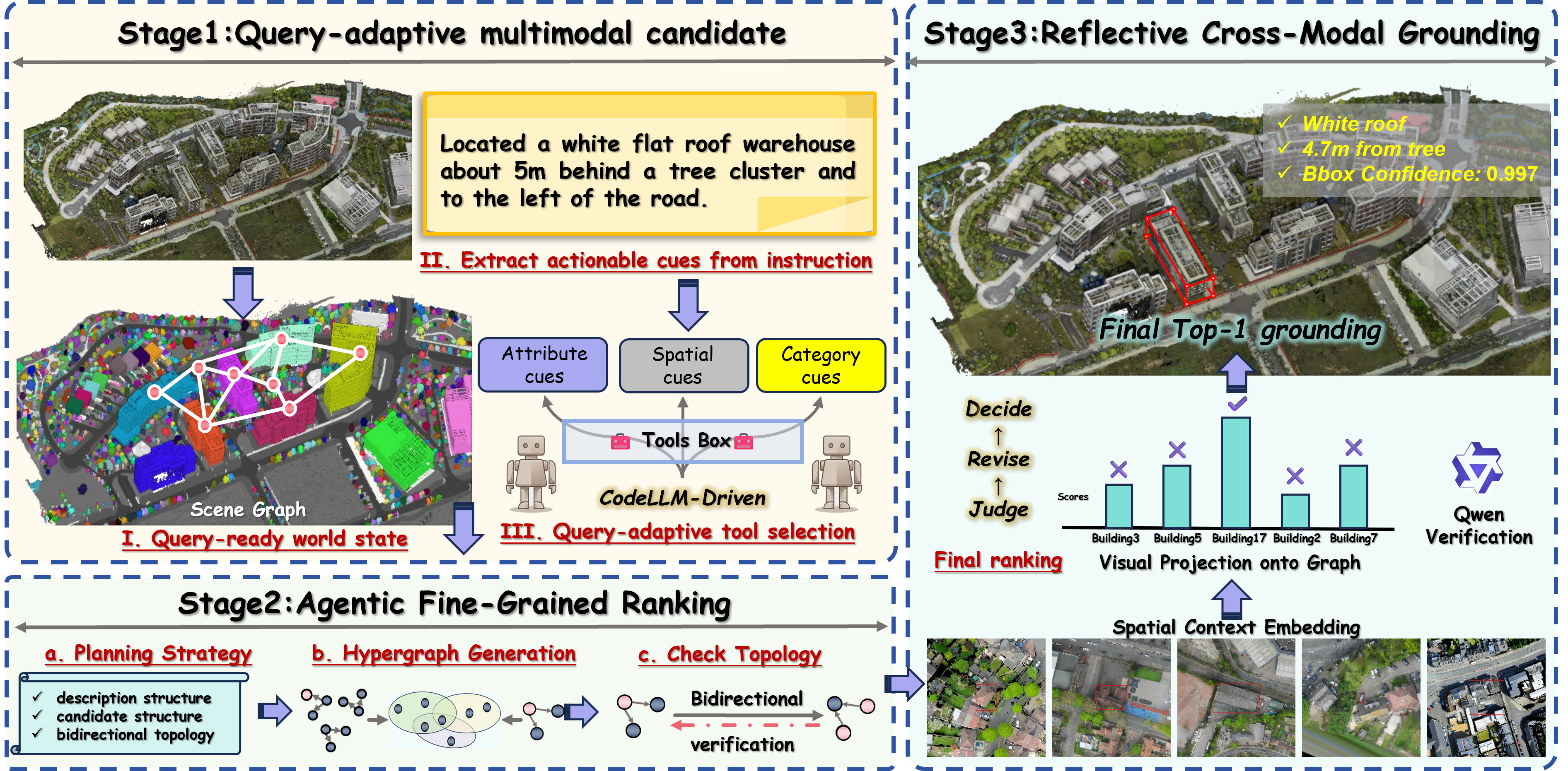}
    \vspace{-6pt}
    \caption{\textbf{Overview of CitySTAR}. The framework first constructs a query-ready world state and performs CodeLLM-driven multimodal candidate grounding, then applies topology-aware graph alignment for target-context verification, and finally combines topology consistency with candidate-centered 2D visual evidence to produce the final grounding result.}
    \vspace{-6pt}
    \label{fig:citystar}
\end{figure}

\section{Method}
\label{Method}

\subsection{Overview}

\textbf{Problem Definition.} Given a city-scale 3D scene \(\mathcal{S}\) and a natural-language query \(q\), the goal is to localize the target instance referred to by \(q\). Let \(\mathcal{I}=\{i_1,\dots,i_N\}\) denote the open-vocabulary instance set extracted from \(\mathcal{S}\), and let \(\mathcal{G}^{s}\) be the corresponding scene graph. Urban grounding depends on semantic, attribute, contextual, and spatial constraints. We formulate the task as:
\begin{equation}
i^{*}=\arg\max_{i_k\in\mathcal{I}} 
\mathrm{Score}(i_k,q,\mathcal{G}^{s}),
\end{equation}
where \(\mathrm{Score}(\cdot)\) measures how well an instance satisfies the constraints implied by \(q\). This formulation allows grounding to consider both the properties of individual objects and their relative configurations in the scene.

\textbf{CitySTAR Overview.} As shown in Figure~\ref{fig:citystar}, CitySTAR follows a three-stage graph-based reasoning pipeline. Stage~1 converts raw point clouds into a query-ready scene graph and grounds category, attribute, and spatial cues through dynamic CodeLLM tool orchestration. Stage~2 aligns query-side and candidate-side hypergraphs to verify target-context topology, removing structurally inconsistent candidates. Stage~3 complements 3D topology with candidate-centered 2D visual verification to resolve residual ambiguity and produce the final grounded instance.

\subsection{Stage 1: Query-adaptive Multimodal Candidate Grounding}
\label{sec:state_cue}
\label{sec:codellm_orchestrator}

\begin{figure}[t]
    \centering
    \includegraphics[width=0.9\linewidth]{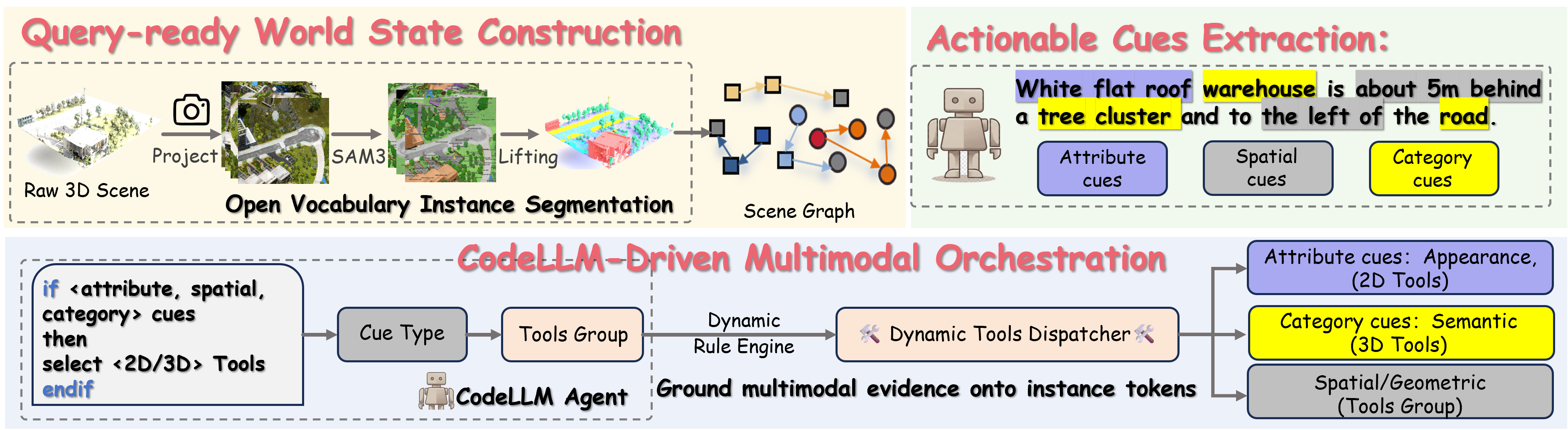}
    \vspace{-6pt}
    \caption{Stage~1 details. CitySTAR builds a query-ready scene graph and populates node and edge evidence with CodeLLM-driven tools to generate a high-recall candidate pool.}
    \vspace{-6pt}
    \label{fig:stage1}
\end{figure}

As illustrated in Figure~\ref{fig:stage1}, Stage~1 transforms the raw urban point cloud into a structured query-ready scene graph. The point cloud is first projected into geo-aware 2D views, where open-vocabulary segmentation identifies candidate objects. The segmented masks are lifted back to 3D and consolidated into instance-level tokens. The resulting state is:
\begin{equation}
\mathcal{G}_{s}=\{\mathcal{I},\mathcal{A},\mathcal{R}\},
\end{equation}
where \(\mathcal{I}\) represents 3D instances, \(\mathcal{A}\) stores semantic and geometric attributes, and \(\mathcal{R}\) encodes spatial relations between instances. This structured representation provides an interpretable and searchable space for subsequent reasoning.

The query \(q\) is organized into actionable cues:
\begin{equation}
\mathcal{C}(q)=\{c_{cat},c_{attr},c_{sp}\},
\end{equation}
where category cues $c_{cat}$ specify the target type, attribute cues $c_{attr}$ indicate visual or geometric properties, and spatial cues $c_{sp}$ encode relative positions with respect to contextual entities. A CodeLLM converts these cues into an executable tool-use plan
\begin{equation}
\mathcal{T}(q)=\{t_1,\dots,t_M\}.
\end{equation}
Each tool evaluates node or edge evidence, producing:
\begin{equation}
R(i\mid q)=\sum_{m=1}^{M} w_m(q)\,s_m(i\mid t_m,\mathcal{G}_{s}),
\end{equation}
where $s_m(\cdot)$ measures the consistency score between instance $i$ and the $m$-th tool's cue, and $w_m(q)$ denotes the query-dependent importance weight of that tool. The output is a high-recall candidate pool annotated with multimodal evidence, which supports Stage~2 reasoning.

\subsection{Stage 2: Topology-aware Soft Graph Alignment}
\label{sec:topology_alignment}
\label{sec:paired_hypergraph}
\label{sec:topology_verifier}

\begin{figure}[t]
    \centering
    \includegraphics[width=0.9\linewidth]{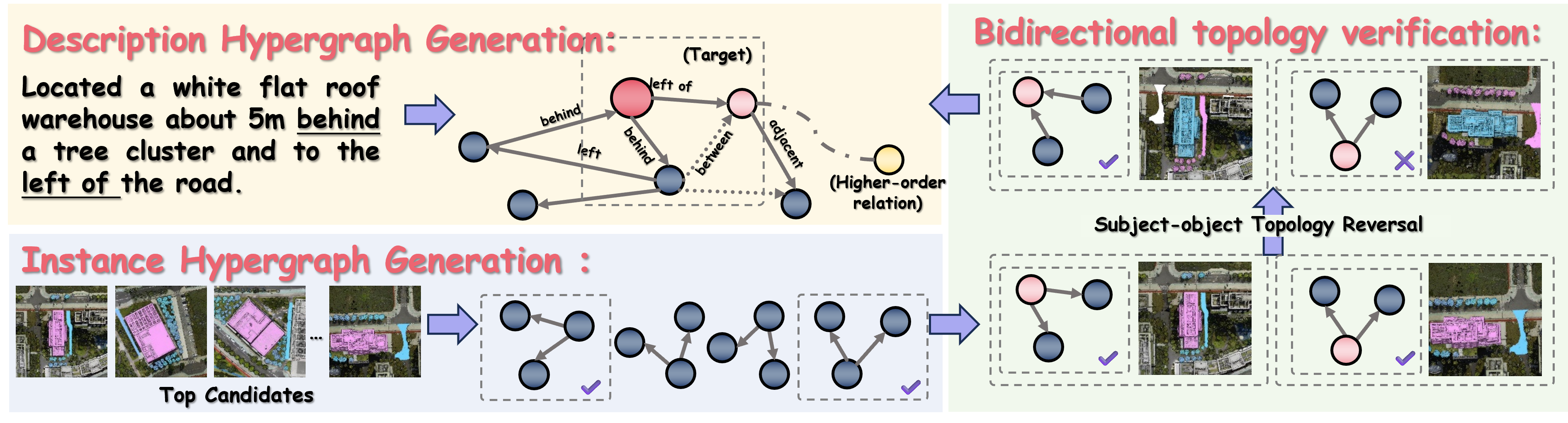}
    \vspace{-6pt}
    \caption{Stage~2 details. Query-side and candidate-side hypergraphs are constructed and aligned to verify target-context topology through bidirectional relation checking.}
    \vspace{-6pt}
    \label{fig:stage2}
\end{figure}

As shown in Figure~\ref{fig:stage2}, Stage~2 shifts the focus from semantic similarity to structural verification. This step is motivated by a common ambiguity in city-scale scenes: multiple candidates may share similar categories, appearances, and coarse locations, while only one candidate satisfies the complete target-context topology described by the query. Therefore, CitySTAR explicitly represents both the language description and each candidate neighborhood as graph structures before computing the topology score.

The query is converted into a hypergraph \(\mathcal{G}^{q}=(\mathcal{V}^{q},\mathcal{R}^{q})\), where nodes represent the target and contextual entities, and hyperedges encode spatial or topological relations. The referred target is marked as the anchor node. This representation converts a flat sentence into a structured relational template. For each candidate \(j\), a local scene-side hypergraph \(\mathcal{G}^{s}_{j}=(\mathcal{V}^{s}_{j},\mathcal{E}^{s}_{j})\) is constructed around its 3D box. The candidate instance is treated as the anchor, and nearby instances are included as contextual nodes. To keep the verification query-conditioned, the scene-side hypergraph only keeps relation types that appear in \(\mathcal{R}^{q}\).

Given the paired hypergraphs, node correspondences are softly established using Stage~1 evidence. For each query-side relation, CitySTAR searches for the best compatible scene-side relation according to node semantics and relation type. Its geometric confidence is used as the relation-level consistency. To reduce false positives caused by reversed subject-object roles, each relation is verified together with its inverse predicate, leading to a bidirectional relation score \(s_{\mathrm{bi}}(e)\). The topology score is computed as:
\begin{equation}
\mathrm{Score}_{\mathrm{topo}}(\mathcal{G}^{q},\mathcal{G}^{s}_{j}) 
= 
\frac{1}{|\mathcal{R}^{q}|} 
\sum_{e\in\mathcal{R}^{q}} s_{\mathrm{bi}}(e),
\end{equation}
where higher scores indicate stronger consistency with the target-context structure described by the query. This step filters candidates that are semantically plausible but relationally inconsistent.

\subsection{Stage 3: Reflective Cross-Modal Grounding}
\label{sec:cross_modal_grounding}

After topology-aware alignment, the remaining candidates are structurally plausible but may still differ in fine-grained appearance or local visual context. Stage~3 addresses this residual ambiguity through candidate-centered 2D visual verification. Rather than using 2D evidence as an independent fallback, CitySTAR uses it as a complementary signal with 3D topology construction. This design preserves the interpretability of topology reasoning while allowing the final decision to benefit from richer visual cues.

For each candidate \(j\), we generate a candidate-centered rendering from the 3D scene. The rendering focuses on the candidate and its surrounding context so that visual attributes and nearby objects can be evaluated jointly. A visual-text reranker~\citep{zhang2025qwen3embeddingadvancingtext} then compares the rendering with the query and produces a visual consistency score \(S_{\mathrm{vis}}(j)\). This score captures fine-grained appearance cues such as roof color, object shape, local texture, and visible contextual objects, which may not be fully resolved by 3D topology alone.

The final prediction combines graph-based topology consistency and candidate-centered visual evidence:
\begin{equation}
j^{*} = 
\arg\max_{j} 
\Big(
\mathrm{Score}_{\mathrm{topo}}(\mathcal{G}^{q},\mathcal{G}^{s}_{j}) 
+ 
\lambda_{\mathrm{2D}} S_{\mathrm{vis}}(j)
\Big),
\end{equation}
where \(\lambda_{\mathrm{2D}}\) balances the contribution of 2D visual verification. In this way, CitySTAR first enforces spatial consistency in the 3D graph and then uses cross-modal visual evidence to select the most reliable target among the structurally plausible candidates.
\section{Experiments}

\subsection{Dataset}

We evaluate CitySTAR on three city-scale grounding benchmarks: \textit{CityRefer}~\citep{miyanishi2023cityrefer}, \textit{CityAnchor}~\citep{li2025cityanchor}, and our \textit{CitySTAR-3D}. CityRefer is built on SensatUrban and mainly covers four coarse target categories, including Building, Car, Ground, and Parking. CityAnchor is constructed from STPLS3D and contains 9 urban object categories. In our evaluation, the validation splits contain over 1,000 grounding queries for CityRefer and over 200 queries for CityAnchor.

\textbf{CitySTAR-3D.}
As summarized in Table~\ref{tab:compact_dataset_comparison}, CitySTAR-3D is designed to better evaluate reasoning-oriented open-vocabulary grounding in large urban scenes. It contains 1286 text-object pairs, including 686 samples from SensatUrban and 600 samples from STPLS3D, and covers 15 semantic groups. Compared with CityRefer and CityAnchor, it provides broader semantic coverage, more complete instance annotations, clearer bounding boxes, and more complex descriptions involving attributes, contextual entities, containment relations, neighbor relations, and spatial layouts. These properties make CitySTAR-3D more suitable for evaluating whether a model can understand target-context topology rather than only match object appearance. 


\begin{table*}[h]
\centering
\vspace{-6pt}
\captionsetup{skip=2pt}
\caption{Dataset comparison of CityRefer, CityAnchor, and CitySTAR-3D.}
\label{tab:compact_dataset_comparison}
\renewcommand{\arraystretch}{1.0}
\setlength{\tabcolsep}{4.5pt}
\footnotesize
\resizebox{\linewidth}{!}{
\begin{tabular}{
p{2.7cm}
p{3.2cm}
p{3.5cm}
p{4.2cm}
p{4.0cm}
}
\toprule
\rowcolor{gray!15}
\textbf{Dataset}
& \textbf{Semantic Scope}
& \textbf{Description Type}
& \textbf{Annotation Quality}
& \textbf{Typical Example} \\
\midrule

CityRefer
& 4 coarse categories
& Simple object or landmark reference
& Standard object boxes
& \textit{car near building} \\

\rowcolor{gray!6}
CityAnchor
& 9 urban object categories
& Object prompts with simple spatial context
& Standard object boxes
& \textit{building beside tree} \\

\rowcolor{orange!14}
\textbf{CitySTAR-3D}
& \textbf{15 semantic groups with fine-grained types}
& \textbf{Attribute, containment, neighbor, and relation-aware descriptions}
& \textbf{Refined boxes with more complete object extents}
& \textit{warehouse west of the river and adjacent to parked cars} \\

\bottomrule
\end{tabular}
}
\vspace{-10pt}
\end{table*}
\subsection{Implementation Details}

All experiments are conducted on a single NVIDIA A800 GPU. We locally deploy a 9B Qwen-3.5 model for cues organization and CodeLLM-driven tool use and  a Qwen3-VL-Reranker-8B model model for visual-based verification. The tool library includes functions for color analysis, 2D appearance matching, 3D semantic retrieval, object size measurement, and inter-object distance computation. More details are provided in the supplementary material.

We evaluate grounding results using IoU and report Acc@0.25 and Acc@0.50. We follow the standard \textit{Novel Objects} (NO) and \textit{Novel Descriptions} (ND) settings, where NO evaluates unseen target instances and ND evaluates new language descriptions for known targets. CitySTAR performs grounding over the full instance space of each city-scale scene and dynamically generates a candidate pool for topology verification and final decision.
\subsection{Performance Evaluation}

Table~\ref{tab:results} reports the quantitative comparison on CityRefer and CityAnchor under the Novel Objects (NO) and Novel Description (ND) settings. CitySTAR consistently outperforms all baselines across both datasets. Compared with the strongest baseline CityAnchor, CitySTAR improves Acc@0.25/Acc@0.50 by \gain{21.85}/\gain{24.82} on CityRefer-NO and by \gain{29.50}/\gain{30.31} on CityRefer-ND. Similar gains are observed on CityAnchor, where CitySTAR improves Acc@0.25/Acc@0.50 by \gain{21.37}/\gain{25.96} under NO and by \gain{8.71}/\gain{9.64} under ND. The larger improvement on ND indicates that CitySTAR is particularly effective for compositional language expressions, where direct object-level matching becomes unreliable.

\begin{figure}[h]
    \centering
    \includegraphics[width=0.9\linewidth]{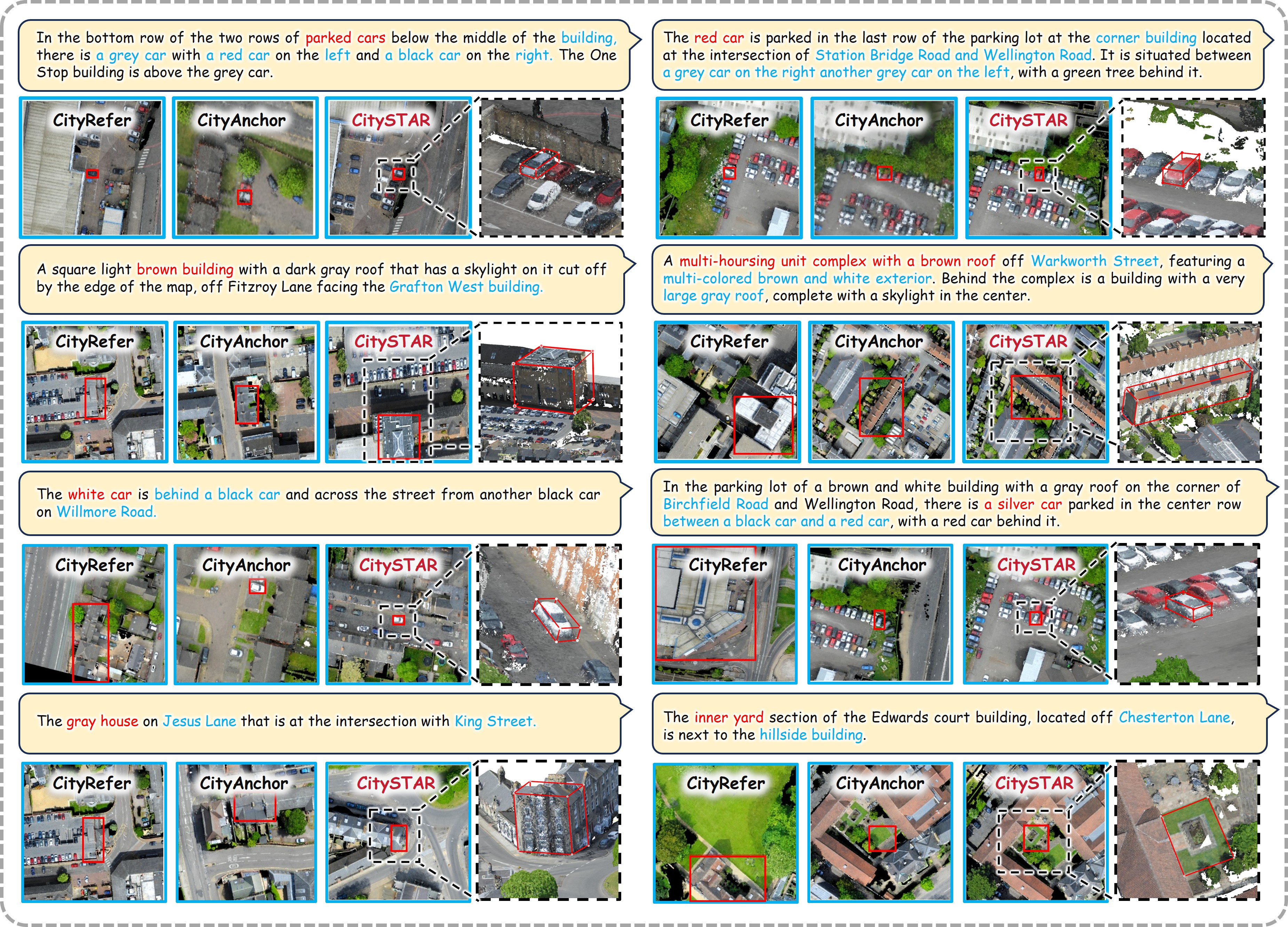}
    \vspace{-6pt}
    \caption{Qualitative grounding comparison on the CityRefer dataset.  Compared with CityRefer and CityAnchor, CitySTAR better handles challenging urban descriptions that involve similar objects, contextual cues and spatial relations.  It reduces wrong-target predictions and improves bounding-box accuracy in complex city-scale scenes.
}
\vspace{-6pt}
    \label{fig:result1}
\end{figure}

\begin{figure}[h]
    \centering
    \includegraphics[width=0.9\linewidth]{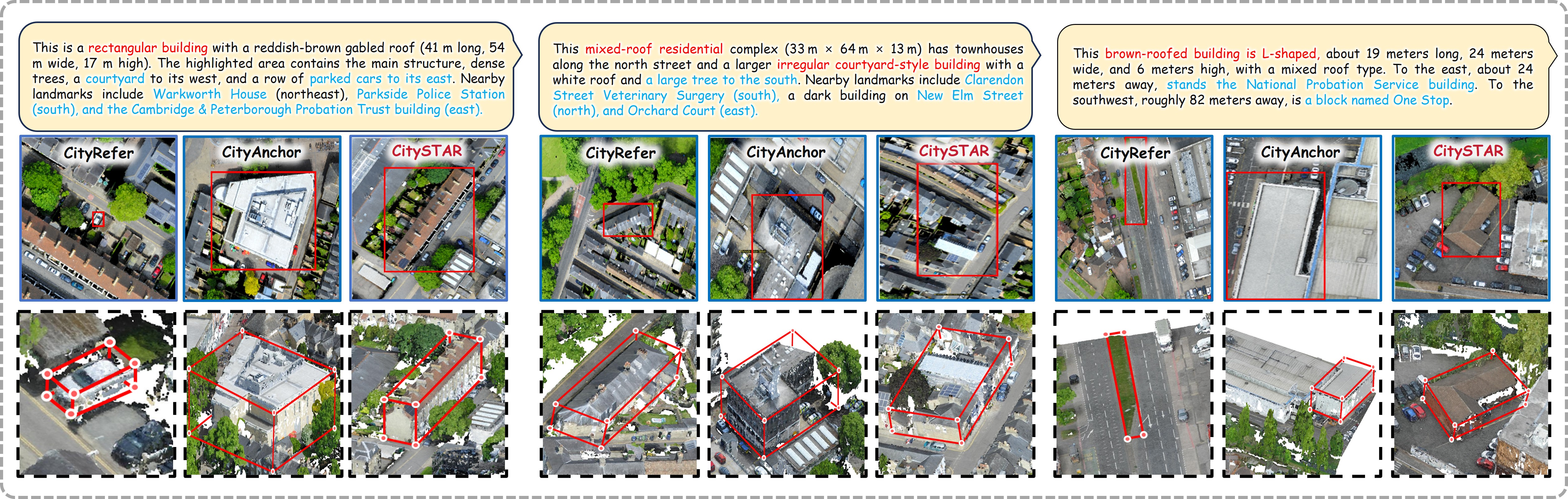}
    \vspace{-6pt}
    \caption{Qualitative grounding comparison on the CitySTAR-3D dataset.  Compared with CityRefer and CityAnchor, CitySTAR demonstrates superior handling of complex urban scenes, including challenging object co-occurrences, contextual cues, and spatial relations.  It reduces wrong-target predictions and improves bounding-box accuracy at city scale.}
    \vspace{-6pt}
    \label{fig:CitySTAR3D_result}
\end{figure}

\begin{figure}[h]
    \centering
    \includegraphics[width=0.9\linewidth]{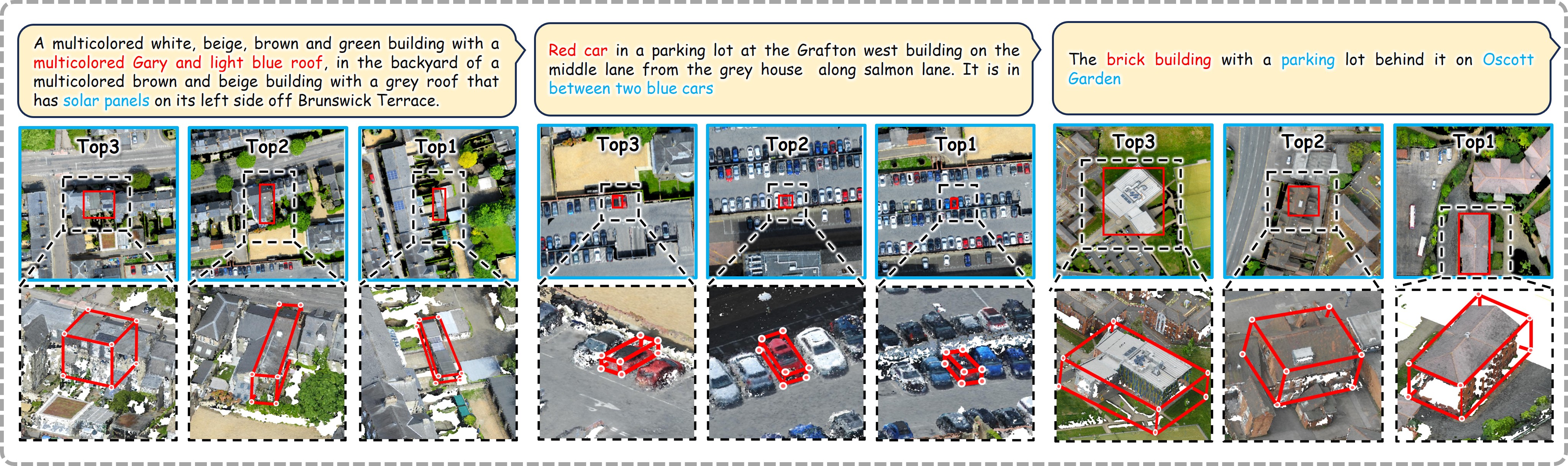}
    \vspace{-6pt}
    \caption{Text-aware Top-3 Candidate Ranking. CitySTAR ranks candidates by partial-to-complete alignment with the query rather than random visual similarity.}
    \vspace{-8pt}
    \label{fig:top3}
\end{figure}
\begin{table}[h]
  \centering
  \caption{Quantitative comparison on CityRefer and CityAnchor datasets.  NO and ND denote Novel Objects and Novel Descriptions.}
  \label{tab:results}
  \scriptsize
  \renewcommand{\arraystretch}{1}
  \resizebox{\linewidth}{!}{%
  \begin{tabular}{lcccccccc}
    \toprule
    \textbf{Method}
   & \multicolumn{2}{c}{\textbf{CityRefer-NO}} 
      & \multicolumn{2}{c}{\textbf{CityRefer-ND}} 
      & \multicolumn{2}{c}{\textbf{CityAnchor-NO}} 
      & \multicolumn{2}{c}{\textbf{CityAnchor-ND}} \\
    \cmidrule(lr){2-3}
    \cmidrule(lr){4-5}
    \cmidrule(lr){6-7}
    \cmidrule(lr){8-9}
    & \textbf{Acc@0.25} & \textbf{Acc@0.50}
    & \textbf{Acc@0.25} & \textbf{Acc@0.50}
    & \textbf{Acc@0.25} & \textbf{Acc@0.50}
    & \textbf{Acc@0.25} & \textbf{Acc@0.50} \\
    \midrule
    InstanceRefer
    & 4.09 & 3.64 & 1.93 & 1.76 & 1.58 & 1.35 & 3.04 & 2.31 \\
    3DVG-Transformer
    & 7.73 & 5.69 & 9.64 & 8.12 & 4.16 & 2.38 & 6.25 & 4.17 \\
    EDA
    & 6.96 & 5.53 & 8.39 & 5.84 & 5.15 & 3.09 & 7.14 & 4.29 \\
    CityRefer
    & 8.34 & 7.47 & 5.07 & 3.49 & 5.73 & 4.16 & 6.07 & 3.95 \\
    \rowcolor{gray!12}
    CityAnchor
    & 50.69 & 46.86 & 53.17 & 50.37 & 41.23 & 35.11 & 47.81 & 43.40 \\
    \midrule
    \rowcolor{orange!18}
    \textbf{CitySTAR (Ours)}
    & \textbf{72.54} & \textbf{71.68}
    & \textbf{82.67} & \textbf{80.68}
    & \textbf{62.60} & \textbf{61.07} & \textbf{56.52} & \textbf{53.04} \\
    \bottomrule
  \end{tabular}%
  }
\end{table}

The gap between Acc@0.25 and Acc@0.50 further reflects localization quality. On CityRefer, CitySTAR drops only \drop{0.86} under NO and \drop{1.99} under ND when the IoU threshold increases from 0.25 to 0.50. In contrast, CityAnchor drops by \drop{3.83} and \drop{2.80}. This smaller degradation suggests that CitySTAR not only retrieves the correct semantic target, but also predicts boxes that better match the object extent. Figure~\ref{fig:result1} supports this observation. CitySTAR reduces wrong-target predictions in scenes with similar objects and produces tighter boxes when the target is determined by contextual and spatial cues.

\begin{table*}[!h]
\centering
\setlength{\tabcolsep}{2.2pt}
\renewcommand{\arraystretch}{1.12}
\scriptsize
\caption{Category-wise Acc@0.25 (\%) comparison on CitySTAR-3D. 
The last column reports the macro-average over the 15 semantic groups.}
\label{tab:unified_dataset_comparison}
\resizebox{\textwidth}{!}{
\begin{tabular}{l*{15}{c}c}
\toprule
\textbf{Method}
& \rotatebox{45}{\textbf{Generic}}
& \rotatebox{45}{\textbf{Residential}}
& \rotatebox{45}{\textbf{Office}}
& \rotatebox{45}{\textbf{Vehicle}}
& \rotatebox{45}{\textbf{Parking}}
& \rotatebox{45}{\textbf{Vegetation}}
& \rotatebox{45}{\textbf{Hospitality}}
& \rotatebox{45}{\textbf{Education}}
& \rotatebox{45}{\textbf{Industrial}}
& \rotatebox{45}{\textbf{Religious}}
& \rotatebox{45}{\textbf{Warehouse}}
& \rotatebox{45}{\textbf{Civic}}
& \rotatebox{45}{\textbf{Pedestrian}}
& \rotatebox{45}{\textbf{Rail}}
& \rotatebox{45}{\textbf{Water}}
& \textbf{Avg.} \\
\midrule
CityRefer 
& 18.75 & 23.30 & 29.03 & 15.15 & 37.21 & 12.50 & 30.56 & 22.86 & 23.81 & 50.00 & 6.25 & 56.25 & 25.00 & 0.00 & 0.00 
& 23.38 \\

\rowcolor{gray!12}
CityAnchor 
& 17.61 & 25.24 & 26.88 & 13.64 & 39.53 & 15.00 & 27.78 & 25.71 & 19.05 & 44.44 & 12.50 & 50.00 & 18.75 & 0.00 & 0.00 
& 22.41 \\

\midrule
\rowcolor{orange!18}
\textbf{CitySTAR (Ours)} 
& \textbf{59.09} & \textbf{63.11} & \textbf{61.29} & \textbf{57.58} & \textbf{67.44} 
& \textbf{45.00} & \textbf{66.67} & \textbf{60.00} & \textbf{61.90} & \textbf{61.11} 
& \textbf{43.75} & \textbf{75.00} & \textbf{31.25} & \textbf{50.00} & \textbf{50.00} 
& \textbf{56.88} \\
\bottomrule
\end{tabular}
}
\end{table*}

Table~\ref{tab:unified_dataset_comparison} further evaluates open-vocabulary grounding on CitySTAR-3D. CitySTAR achieves the best performance across all 15 semantic groups, with an average category accuracy of 56.88\%, compared with 23.38\% for CityRefer and 22.41\% for CityAnchor. The improvement is notable on functional building categories such as Warehouse, Industrial, Office, Residential, and Hospitality, where semantic labels alone are often insufficient. CitySTAR also obtains non-zero results on Rail and Water, while both baselines fail. Figure~\ref{fig:CitySTAR3D_result} further shows qualitative comparisons on CitySTAR-3D, illustrating more accurate grounding in complex urban scenes. Figure~\ref{fig:top3} shows that top-ranked candidates are text-relevant rather than random, reflecting semantically meaningful ranking.

\begin{table}[!h]
  \centering
  \caption{Ablation study on the CityRefer dataset.}
  \label{tab:ablation}
  \scriptsize
  \setlength{\tabcolsep}{12pt}
  \renewcommand{\arraystretch}{0.86}
  \resizebox{\linewidth}{!}{%
  \begin{tabular}{lcccc}
    \toprule
    \textbf{Variant}
    & \multicolumn{2}{c}{\textbf{CityRefer-NO}}
    & \multicolumn{2}{c}{\textbf{CityRefer-ND}} \\
    \cmidrule(lr){2-3}
    \cmidrule(lr){4-5}
    & \textbf{Acc@0.25} & \textbf{Acc@0.50}
    & \textbf{Acc@0.25} & \textbf{Acc@0.50} \\
    \midrule

    \rowcolor{gray!10}
    \multicolumn{5}{l}{\textit{Building stage}} \\
    Replace SAM3 with SoftGroup++
    & 44.91 & 43.70 & 48.01 & 46.22 \\

    \midrule
    \rowcolor{gray!10}
    \multicolumn{5}{l}{\textit{Stage 1: CodeLLM-driven candidate grounding}} \\
    w/o spatial query
    & 37.13 & 34.54 & 43.23 & 41.83 \\
    w/o 2D query
    & 38.34 & 35.92 & 43.63 & 42.43 \\
    Full tool use with Qwen3-Coder
    & 45.42 & 43.18 & 46.02 & 44.02 \\
    Full tool use with Qwen3.5-9B
    & 45.42 & 43.52 & 46.63 & 44.42 \\

    \midrule
    \rowcolor{gray!10}
    \multicolumn{5}{l}{\textit{Stage 1 + Stage 2: topology-aware verification}} \\
    + Node-level hypergraph matching
    & 51.81 & 49.22 & 56.18 & 53.59 \\
    + Node + edge topology verification
    & 65.11 & 63.73 & 72.31 & 70.72 \\

    \midrule
    \rowcolor{gray!10}
    \multicolumn{5}{l}{\textit{Effect of removing topology verification}} \\
    w/o Stage 2
    & 56.13 & 53.36 & 52.78 & 51.79 \\

    \midrule
    \rowcolor{gray!10}
    \multicolumn{5}{l}{\textit{Stage 1 + Stage 2 + Stage 3: final visual verification}} \\
    + LLaVA visual verification
    & 68.39 & 66.84 & 76.10 & 74.30 \\
    \rowcolor{orange!18}
    \textbf{+ Qwen3-VL-Reranker-8B verification}
    & \textbf{72.54} & \textbf{71.68}
    & \textbf{82.67} & \textbf{80.68} \\

    \bottomrule
  \end{tabular}%
  }
\end{table}

\subsection{Ablation Study}

Table~\ref{tab:ablation} verifies the contribution of each component. Stage~1 shows that multimodal cue grounding is necessary for reliable candidate generation. Removing spatial queries reduces Acc@0.50 by \drop{8.98} under CityRefer-NO, while removing 2D appearance queries reduces it by \drop{7.60}. This confirms that metric spatial cues and visual appearance cues provide complementary evidence before topology verification. Replacing Qwen3-Coder with Qwen3.5-9B brings only moderate gains of \gain{0.34} and \gain{0.40} on Acc@0.50 under NO and ND, suggesting that the main improvement does not come from the CodeLLM backbone alone.

Stage~2 provides the key gain by introducing explicit target-context reasoning. Adding hypergraph node verification improves Acc@0.50 from 43.52 to 49.22 under NO and from 44.42 to 53.59 under ND. Further adding hypergraph edge verification increases Acc@0.50 to 63.73 and 70.72, yielding additional gains of \gain{14.51} and \gain{17.13}. This demonstrates that edge-level topology is crucial for distinguishing candidates that are visually plausible but relationally inconsistent.

Stage~3 improves the final decision through candidate-centered visual verification. Compared with LLaVA verification directly assigning scores,, the Qwen3-VL-Reranker-8B improves Acc@0.50 by \gain{4.84} under NO and \gain{6.38} under ND. The importance of Stage~2 is further confirmed by the ``w/o Stage~2'' variant, which drops from 71.68 to 53.36 Acc@0.50 under NO and from 80.68 to 51.79 under ND. These drops of \drop{18.32} and \drop{28.89} show that visual verification alone cannot replace explicit 3D topology reasoning. Overall, CitySTAR benefits from the progressive design of cue grounding, topology verification, and cross-modal visual verification. 

\section{Limitation and Conclusion}

\textbf{Limitation.}
Although CitySTAR shows strong performance for city-scale 3D grounding, several aspects can be further improved. First, more accurate open-vocabulary instance extraction may further strengthen the query-ready scene graph in highly cluttered urban scenes. Second, extending the current spatial relation set to finer metric, functional, and commonsense relations could support broader urban reasoning tasks. Third, while our overall inference time is lower than that of the compared methods, CodeLLM-driven tool use and visual-enhanced reranking still contribute additional computational overhead, motivating future work on lightweight tool scheduling, model acceleration, and broader QA task exploration.

\textbf{Conclusion.}
We propose CitySTAR, a training-free framework that brings structured reasoning to city-scale 3D grounding. Rather than treating grounding as direct similarity matching, CitySTAR first builds a query-ready scene graph, then uses CodeLLM-driven multimodal candidate grounding to obtain a compact proposal set, followed by paired hypergraph topology verification to resolve target-context ambiguity. A visual-enhanced reflective reranking stage further verifies the candidates across both real 3D space and visual space. We also develop CitySTAR-3D, an enhanced benchmark with improved instance annotations and more compositional urban descriptions. Results on CityRefer, CityAnchor, and CitySTAR-3D show that CitySTAR achieves more reliable open-world urban grounding, especially for queries that require attribute understanding and spatial reasoning.


\bibliographystyle{plainnat}
\bibliography{main}

\clearpage


\end{document}